\documentclass[times, twoside, watermark]{zHenriquesLab-StyleBioRxiv}
\usepackage{blindtext}
\usepackage{lipsum}
\leadauthor{Yongbing Zhang} 

\begin{document}

\title{Multi tasks RetinaNet for mitosis detection}
\shorttitle{Multi tasks RetinaNet for mitosis detection}

\author[1]{Yang Chen}
\author[2]{Ziyue Wang}
\author[1]{Zijie Fang}
\author[1]{Hao Bian}
\author[2]{Yongbing Zhang}

\affil[1]{Shenzhen International Graduate School, Tsinghua University, China}
\affil[2]{Department of Computer Science, Harbin Institute of Technology, China}

\maketitle

\begin{abstract}
The account of mitotic cells is a key feature in tumor diagnosis. However, due to the variability of mitotic cell morphology, it is a highly challenging task to detect mitotic cells in tumor tissues. At the same time, although advanced deep learning method have achieved great success in cell detection, the performance is often unsatisfactory when tested data from another domain (i.e. the different tumor types and different scanners). Therefore, it is necessary to develop  algorithms for detecting mitotic cells with  robustness in domain shifts scenarios. Our work further proposes a foreground detection and tumor classification task based on the baseline(Retinanet), and 
utilizes data augmentation to improve the domain generalization performance of our model. We achieve the state-of-the-art performance (F1 score: 0.5809) on the challenging premilary test dataset.

\end {abstract}


\begin{corrauthor}
ybzhang08@hit.edu.cn\\
Equal contribution: Yang Chen, Ziyue Wang
\end{corrauthor}

\section*{Introduction}
Histopathology is the gold standard for tumor diagnosis and prognosis, especially with the rapid development of deep learning technology and the popularization of whole slide image (WSI) scanners in recent years, computational pathology has received high attention.

In the clinical diagnosis of tumors, judging the degree of malignant proliferation of tumors is a very key prognostic indicator, and also a key reference for tumor grading and treatment. Usually pathologists need to use KI67 on histopathological sections by immunohistochemistry\cite{cai2021generalizing}. The tissue sections were stained with stains, and the degree of malignancy of tumor proliferation was judged according to the degree of immunohistochemical staining.

However, immunohistochemistry is more time-consuming than conventional HE-stained sections, and the cost is higher in some medically underdeveloped areas. In recent years, with the development of deep learning in the direction of target detection in computer vision, it has become an increasing challenge to identify mitotic cells directly from HE-stained histopathological sections by detection algorithm.

\section*{Material and Methods}

\subsection{Dataset description}
Mitosis domain generalization challenge 2022 (MIDOG 2022) \cite{mitosis} is a further expansion of tumor types and types of scanners based on the MIDOG 2021 competition \cite{aubreville2022mitosis,aubreville2022biomedical}. In the MIDOG 2021 competition, the competition only focused on the color generalization error caused by different scanners on the same tumor (breast cancer). In MIDOG 2022, the organizers of the competition digitized different tumor cell sections from different laboratories and different species (i.e. human, canine) under different scanners. MIDOG 2022 aims to explore mitosis cell detection algorithms with multiple 
domains (differences in tissue and collection methods).

The training dataset contains 405 WSIs, including 6 tumor tissue types and 2 species: Canine Lung Cancer, Human Breast Cancer, Canine Lymphoma, Human Neuroendocrine Tumor, Canine Cutaneous Mast Cell Tumor, and Human Melanoma; The test dataset contains 10 kinds of tumor tissues and the type is unknown.

The mitotic cells of the competition are determined by visual assessment by a trained pathologist. The competition organizer provided mitotic cell annotations for the five tumor tissue sections in the training dataset, excluding Human melanoma, and the corresponding scanner type for each tumor type. Since breast cancer contains 3 scanner types at the same time, and there is no specific scanner type corresponding to each WSI, we consider that the scanner type of breast cancer tissue section is unknown.

\subsection{Methods}
We propose a multi-task mitotic cell detection model based on RetinaNet, in which the main improvements include three parts:
\begin{itemize}
    \item An auxiliary classification network is used to classify tumors, including six categories: prostate cancer, lymphoma, lung cancer, melanoma, breast cancer and mast cell tumor.
    \item An auxiliary classification network is used to classify whether the patch contains mitotic cells (or hard samples).
    \item A data augmentation transform is used to improve the domain generalization ability of the model and detect mitotic cells of different types of tumors (different scanners).
\end{itemize}

\subsection{Multi task auxiliary classification}
\begin{figure}
\centering
\includegraphics[width=.8\linewidth]{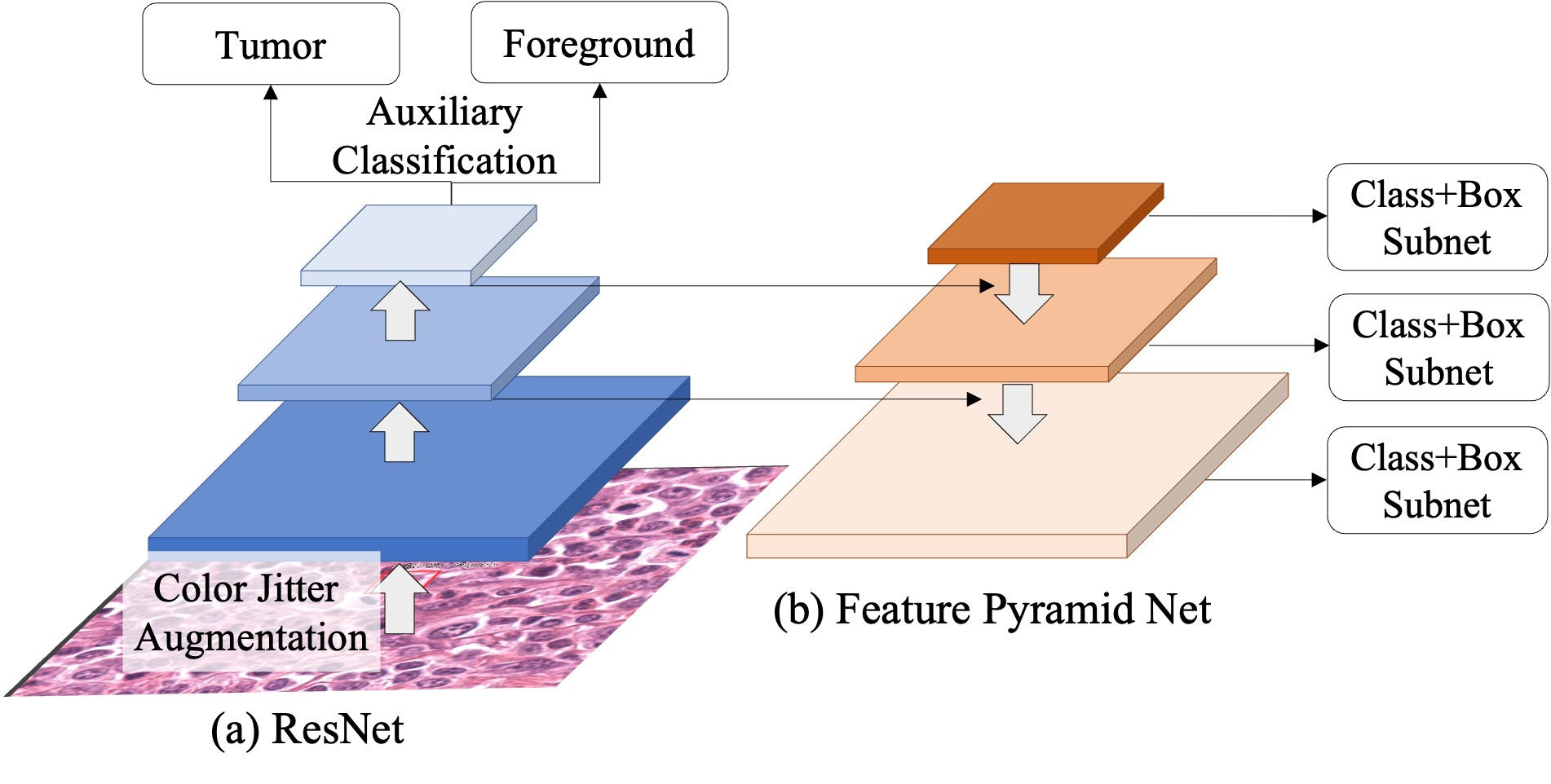}
\caption{Overview of multi task RetinaNet.}
\label{fig:overview}
\end{figure}

As shown in the Fig.\ref{fig:overview}, we firstly select the deep feature of FPN in RetinanNet \cite{lin2020focal} and then add two fully connected network auxiliary classification tasks after this deep feature.

For the tumor auxiliary classification task, we use cross entropy. During the training, we consider 6 tissue types at the same time. The loss function is defined as follows:
\begin{equation}
CE \left(p_{c}\right)= -\sum_{i=0}^{C-1} y_{i} \log \left(p_{i}\right)=-\log \left(p_{c}\right),
\end{equation}
where, $y_{i}$ represents the real tumor category of the ith object, $p_{i}$ indicates the predicted tumor category probability, and $C$ indicates the number of tumor categories.

For the foreground auxiliary classification task, we firstly regard the patches containing mitotic cells or hard samples as foreground classes, and the patches that do not contain mitotic cells or hard samples as background classes, and use Focal loss as the loss function, which is defined as follows :
\begin{equation}
FL \left(p_{c}\right)=-\alpha_{c}\left(1-p_{c}\right)^{\gamma} \log \left(p_{c}\right),
\end{equation}
Where $\alpha$ and $\gamma$ are two super parameters of focal loss, which are used to adjust the contribution of difficult samples to classification.

\subsection{Data Augmentation}
We use the Color Jitter data enhancement module to improve the domain generalization ability of the model, where the parameter settings of Color Jitter are: brightness = 0.35, contrast = 0.2, saturation=0.1, hue=0.1. Other data augmentation includes: random crop, random horizontal flip and vertical flip.
\subsection{Hyperparameter setting}
We use Adam optimizer. The learning rate is 1e-5, and the batch is 16. For RetinaNet's FPN, we use the ResNet 50 network based on ImageNet 1K pre-training. For the hyperparameter of Focus loss, we set $\alpha$ = 0.25 and $\gamma$ = 2.

\section*{Results}
In order to verify the mitotic cell detection performance and generalization performance of our proposed model, we re-divided the training set and validation set according to the original training set. The newly constructed training set includes four tumor tissues: canine lung cancer, human breast cancer, canine lymphoma and canine cutaneous mastocytoma., The test set includes human neuroendocrine tumor.
Under the general hyperparametric setting, we verified the performance of multi task RetinaNet through ablation studies.  

\subsection{Ablation study}
As show in Tabel~\ref{tab:ablation}, the basic performence of RetinaNet's mAP (Iou is 0.5) on mitosis cell detection is 0.433. Compared with the baseline, the three components proposed by us on this basis can improve the detection performance.The most obvious performance improvement is foreground auxiliary classification component. When we integrate the three proposed components, the performance of the model reaches the maximum 0.486 mAP.

\begin{table}[h]
\caption{Ablation studys on Multi task RetinaNet.}
\begin{tabular}{|c|c|c|c|}
\hline
\begin{tabular}[c]{@{}c@{}}foreground \\ classification\end{tabular} & \begin{tabular}[c]{@{}c@{}}tumor \\ classification\end{tabular} & \begin{tabular}[c]{@{}c@{}}data \\ augmentation\end{tabular} & mAP(iou=0.5) \\ \hline
                                                                    &                                                                &                                                             & 0.433        \\ \hline
                                                                    &                                                                & \checkmark                                                            & 0.449        \\ \hline
                                                                    & \checkmark                                                               &                                                             & 0.458        \\ \hline
\checkmark                                                                    &                                                                &                                                             & 0.451        \\ \hline
                                                                    & \checkmark                                                               & \checkmark                                                            & 0.463        \\ \hline
\checkmark                                                                    &                                                                & \checkmark                                                            & 0.465        \\ \hline
\checkmark                                                                    & \checkmark                                                               &                                                             & 0.473        \\ \hline
\checkmark                                                                    & \checkmark                                                               & \checkmark                                                            & 0.486        \\ \hline
\end{tabular}
\label{tab:ablation}
\end{table}

\section*{Discussion}

Our analysis shows that for the mitotic cell detection task, adding auxiliary classification loss helps to significantly improve the detection performance of the model. For the domain generalization problem from different scanners, compared with the previous domain adaptation algorithm to solve the problem In order to improve the generalization of the model, we propose that the use of colorjitter data augmentation can also increase the domain generalization performance of the model under different Color disturbance of scanners. Especially when the test images come from unknown, the method may be more robust than domain adversarial algorithms.

\section*{Bibliography}
\bibliography{literature}

\begin{thebibliography}{5}
\providecommand{\natexlab}[1]{#1}
\providecommand{\url}[1]{\texttt{#1}}
\expandafter\ifx\csname urlstyle\endcsname\relax
  \providecommand{\doi}[1]{doi: #1}\else
  \providecommand{\doi}{doi: \begingroup \urlstyle{rm}\Url}\fi

\bibitem[Cai et~al.()Cai, Zhu, Cui, Li, Wu, Zhang, and
  Yang]{cai2021generalizing}
Jiatong Cai, Chenglu Zhu, Can Cui, Honglin Li, Tong Wu, Shichuan Zhang, and Lin
  Yang.
\newblock Generalizing {{Nucleus Recognition Model}} in {{Multi-source Ki67
  Immunohistochemistry Stained Images}} via {{Domain-Specific Pruning}}.
\newblock In Marleen de~Bruijne, Philippe~C. Cattin, Stéphane Cotin, Nicolas
  Padoy, Stefanie Speidel, Yefeng Zheng, and Caroline Essert, editors,
  \emph{Medical {{Image Computing}} and {{Computer Assisted Intervention}} –
  {{MICCAI}} 2021}, Lecture {{Notes}} in {{Computer Science}}, pages 277--287.
  {Springer International Publishing}.
\newblock ISBN 978-3-030-87237-3.
\newblock \doi{10/gndqfw}.

\bibitem[mit()]{mitosis}
{{MItosis DOmain Generalization Challenge}} 2022 | {{Zenodo}}.
\newblock https://zenodo.org/record/6362337.

\bibitem[Aubreville et~al.(2022{\natexlab{a}})Aubreville, Stathonikos, Bertram,
  Klopleisch, {ter Hoeve}, Ciompi, Wilm, Marzahl, Donovan, Maier, Breen,
  Ravikumar, Chung, Park, Nateghi, Pourakpour, Fick, Hadj, Jahanifar, Rajpoot,
  Dexl, Wittenberg, Kondo, Lafarge, Koelzer, Liang, Wang, Long, Liu, Razavi,
  Khademi, Yang, Wang, Veta, and Breininger]{aubreville2022mitosis}
Marc Aubreville, Nikolas Stathonikos, Christof~A. Bertram, Robert Klopleisch,
  Natalie {ter Hoeve}, Francesco Ciompi, Frauke Wilm, Christian Marzahl,
  Taryn~A. Donovan, Andreas Maier, Jack Breen, Nishant Ravikumar, Youjin Chung,
  Jinah Park, Ramin Nateghi, Fattaneh Pourakpour, Rutger H.~J. Fick, Saima~Ben
  Hadj, Mostafa Jahanifar, Nasir Rajpoot, Jakob Dexl, Thomas Wittenberg,
  Satoshi Kondo, Maxime~W. Lafarge, Viktor~H. Koelzer, Jingtang Liang, Yubo
  Wang, Xi~Long, Jingxin Liu, Salar Razavi, April Khademi, Sen Yang, Xiyue
  Wang, Mitko Veta, and Katharina Breininger.
\newblock Mitosis domain generalization in histopathology images -- {{The
  MIDOG}} challenge, April 2022{\natexlab{a}}.

\bibitem[Aubreville et~al.(2022{\natexlab{b}})Aubreville, Zimmerer, and
  Heinrich]{aubreville2022biomedical}
Marc Aubreville, David Zimmerer, and Mattias Heinrich, editors.
\newblock \emph{Biomedical {{Image Registration}}, {{Domain Generalisation}}
  and {{Out-of-Distribution Analysis}}: {{MICCAI}} 2021 {{Challenges}}:
  {{MIDOG}} 2021, {{MOOD}} 2021, and {{Learn2Reg}} 2021, {{Held}} in
  {{Conjunction}} with {{MICCAI}} 2021, {{Strasbourg}}, {{France}},
  {{September}} 27\textendash{{October}} 1, 2021, {{Proceedings}}}, volume
  13166 of \emph{Lecture {{Notes}} in {{Computer Science}}}.
\newblock {Springer International Publishing}, {Cham}, 2022{\natexlab{b}}.
\newblock ISBN 978-3-030-97280-6 978-3-030-97281-3.
\newblock \doi{10.1007/978-3-030-97281-3}.

\bibitem[Lin et~al.(2020)Lin, Goyal, Girshick, He, and Dollar]{lin2020focal}
Tsung-Yi Lin, Priya Goyal, Ross Girshick, Kaiming He, and Piotr Dollar.
\newblock Focal {{Loss}} for {{Dense Object Detection}}.
\newblock \emph{IEEE Transactions on Pattern Analysis and Machine
  Intelligence}, 42\penalty0 (2):\penalty0 318--327, February 2020.
\newblock ISSN 0162-8828, 2160-9292, 1939-3539.
\newblock \doi{10/gft2dc}.

\end{thebibliography}

\end{document}